\documentclass{article}

\usepackage[final]{corl_2017} 

\usepackage[utf8]{inputenc} 
\usepackage[T1]{fontenc}    
\usepackage{hyperref}       
\usepackage{url}            
\usepackage{booktabs}       
\usepackage{amsfonts}       
\usepackage{nicefrac}       
\usepackage{microtype}      

\usepackage{amsmath}
\usepackage{wrapfig}
\usepackage{stackengine}
\usepackage{mwhmacros}

\title{The Intentional Unintentional Agent: \\Learning to Solve Many Continuous Control Tasks Simultaneously}

\author{
 Serkan Cabi \quad Sergio G\'omez Colmenarejo \quad Matthew W.\ Hoffman \\  \textbf{ Misha Denil \quad Ziyu Wang \quad Nando de Freitas} \\
  DeepMind \\
  \texttt{\{cabi,sergomez,mwhoffman,mdenil,ziyu,nandodefreitas\}@google.com}
}

\begin{document}
\maketitle

\begin{abstract}
This paper introduces the Intentional Unintentional (IU) agent. This agent endows the deep deterministic policy gradients (DDPG) agent for continuous control with the ability to solve several tasks simultaneously. Learning to solve many tasks simultaneously has been a long-standing, core goal of artificial intelligence, inspired by infant development and motivated by the desire to build flexible robot manipulators capable of many diverse behaviours. We show that the IU agent not only learns to solve many tasks simultaneously but it also learns faster than agents that target a single task at-a-time. In some cases, where the single task DDPG method completely fails, the IU agent successfully solves the task. To demonstrate this, we build a playroom environment using the MuJoCo physics engine, and introduce a grounded formal language to automatically generate tasks.
\end{abstract}


\section{Introduction}
\label{sec:intro}

Imagine a toddler in a playroom trying to bring two blocks together. While purposely focusing on this task, the infant is accomplishing many other goals incidentally both simpler and more complex: gazing, extending the arms, sitting, bending sideways, reaching, grasping, navigating around obstacles,  dodging a looming object thrown by a sibling, sensing texture, sensing temperature, and so on. Over the first year, infants also display a wide range of spontaneous movements: kicks, stomps, sways, flaps, flails, rocks, rubs, nods, shakes, bounces, bangs, waves, wiggles and so on \citep{thelen:1979,adolph:2015}.  

We hypothesize that a single stream of experience offers agents the opportunity to learn and perfect many policies both on purpose and incidentally, thus accelerating the acquisition of grounded knowledge.

To investigate this hypothesis, we propose a deep actor-critic architecture, trained with deterministic policy gradients \citep{silver:2014,lillicrap:2016}, for learning several policies concurrently. The architecture enables the agent to attend to one task on-policy, while unintentionally learning to solve many other tasks off-policy. Importantly, the policies learned unintentionally can be harnessed for intentional use even if those policies were never followed before.

\begin{figure}
    \centering
    \includegraphics[width=0.2\linewidth]{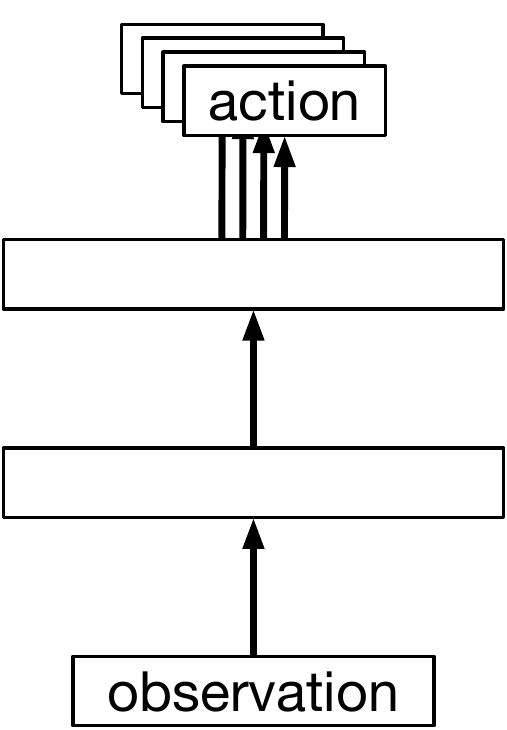}
    \hspace{1em}
    \includegraphics[width=0.27\linewidth]{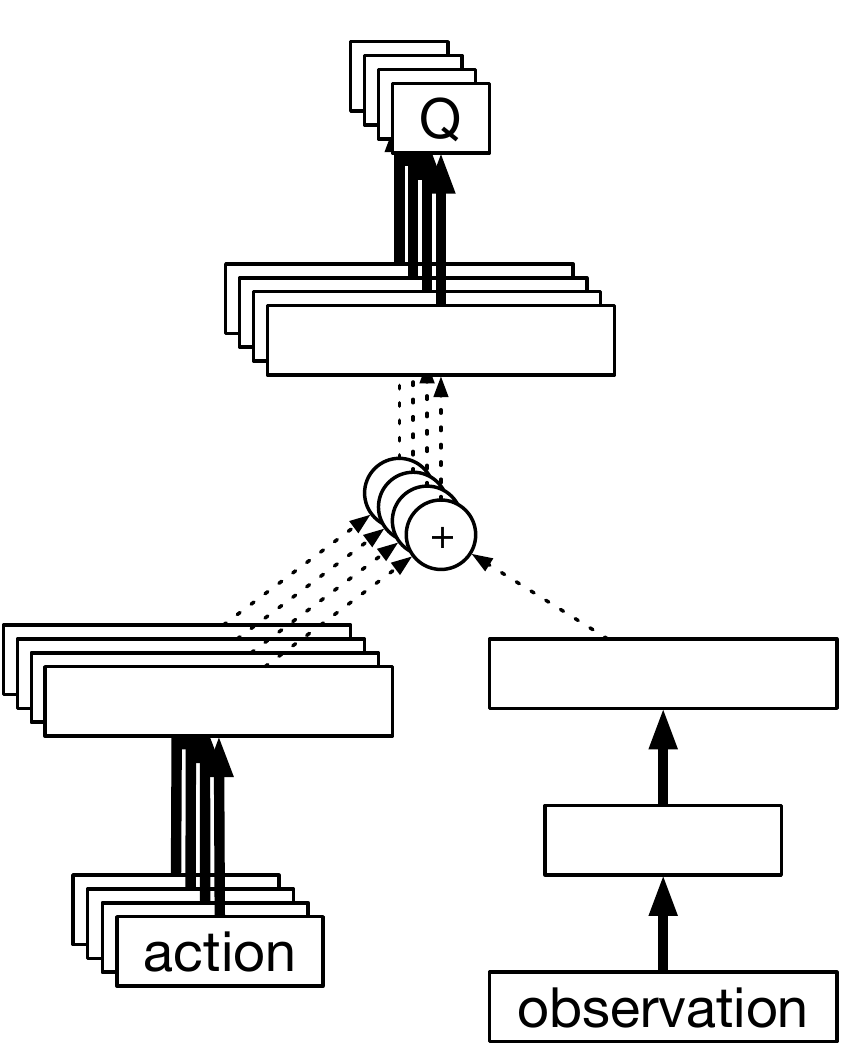}
    \caption{The IU architecture. The actor network on the left consists of two shared MLP-tanh layers followed by non-shared MLP-tanh layers to produce the multivariate actions for each policy (4 policies in this diagram). The right hand side shows the critic network. The action vectors provided by the policies are fed into a non-shared MLP-tanh layer, which is then point-wise added to the ouput of a two layer MLP-tanh network applied to the observation. The resulting activations are processed by non-shared linear layers to produce the $Q$ values.}
    \label{fig:actor-critic-nets}
\end{figure}

More precisely, this {intentional-unintentional architecture}, shown in Figure~\ref{fig:actor-critic-nets},  consists of two neural networks. The actor neural network has multiple-heads representing different policies with shared lower-level representations. The critic network represents several state-action value functions, sharing a common representation for the observations.  

The architecture alone does not suffice for investigating our hypothesis. We also need a flexible way of generating many diverse tasks and a suitable environment. To this end, we introduce an automatic procedure to generate semantic goals for the agent. We also introduce a physical environment, with gravity, a ground, rigid objects, and a simple embodied agent. While some characteristics of the physical world change with experiences, the laws of physics and the body remain fixed to allow for transfer and continual learning.  

While being specific, this study aims to address generality in artificial intelligence by designing agents capable of doing many things, to overcome the problem of sparse rewards associated with conventional reinforcement learning by generating and controlling a stream of reward functions. We design embodied agents situated in a physical environment resembling a playroom, not only as a consequence of being inspired by infant development, but also as a result of being interested in eventual transfer to the world of flexible manufacturing with robots.

\subsection{Related work}

Thinking about an agent as immersed in a stream of multivariate rewards provides us with a powerful alternative to the conventional univariate reward reinforcement learning framework.

Recently, \citet{Jaderberg:2017} consider Asynchronous Advantage Actor Critic (A3C) agents \citep{mnih:2016} immersed in a sensorimotor stream. The agents are designed to achieve an extrinsic scalar reward, but are endowed with  \emph{auxiliary control tasks} and  \emph{auxiliary predictive tasks}. The auxiliary control tasks (pixel changes and simple network features) are shown to enable the A3C agent to learn to achieve the scalar reward faster in domains where the action-space is discrete. This paper will show more dramatic gains for continuous action spaces. 

We refer to the task whose behavior the agent follows during training as the {\it intentional} task, and to the remaining tasks as {\it unintentional}. Philosophically, in our work, the unintentional tasks are not thought of as playing a mere auxiliary role, but they can themselves become the intentional task. Our tasks are of a semantic nature, for example ``{\it move the red block east of the blue block}'', and hence it is sensible to learn a library of such tasks for potential future re-use.

\citet{sutton:2011} introduced the horde architecture to learn grounded knowledge from an unsupervised sensorimotor stream. While not focusing on the issue of representation, with the value functions being trained separately with different weights, the Horde provides much inspiration for this paper.

The concept of general value functions introduced in the horde architecture was further explored by \citet{Schaul:2015}, and has connections with research on options, successor representations and hierarchical RL \citep{Dietterich:1998,Sutton:1998,Kulkarni:2016}. At the time of writing this paper, \citet{vanSeijen2017} introduced a linear decomposition of reward functions, related to \citep{russell2003}, whereby several action-value functions are learned separately with Deep-Q-Networks (DQN) \citep{Mnih:2015}.

Multivariate reward feedback appears in sequential multi-objective decision making \citep{Roijers:2013}, in predictive decision making for 3D games \citep{Dosovitskiy:2017}, and in agents that use auxiliary predictive tasks to reduce sample complexity \citep{li:2015,lample:2016}. 

Our work is related to learning neural networks with a static or adaptive curricula \citep{Bengio:2009,Zaremba:2014,reed:2015}, and to learning curricula for training neural networks using bandit techniques and Bayesian optimization \citep{Tsvetkov:2016,Graves:2017}. 

In developmental psychology, there are many studies on incidental activity and its consequences on motor development, including fetal and neonatal movement \citep{sparling:1999}, twitching during sleep \citep{blumberg:2013}, stereotypies and flails in infants, and gross motor play \citep{adolph:2015}. 

Findings in developmental psychology have inspired the design or robotic systems in the field of developmental robotics \citep{lungarella:2003}. It has been also been argued that robots provide a platform for examining many of these findings. We believe that rich physical simulators are a viable alternative, provided that we keep advancing environments and task generation mechanisms as done in this paper. The importance of bodies and physical environments in the study of artificial intelligence has been championed by many, notably by \citet{brooks:1991}. 

The automatic construction of grounded reward functions with formal languages has become a topic of great interest
in recent months \citep{Littman2017a,Yu2017,Hermann2017,Denil2017}.

\section{The Intentional Unintentional Agent}

Policy gradient algorithms form a very popular, if not the most popular, class of continuous action reinforcement learning algorithms. The fundamental basis of many of these algorithms is the \emph{policy gradient theorem} \citep{sutton:1999}. However, this approach necessitates the use of stochastic policies which can complicate the process of learning off-policy. More recently, a {deterministic policy gradient theorem} has been formulated by \citep{silver:2014} which removes this need. This approach was later extended in \citep{lillicrap:2016} to modern deep neural network actor-critic architectures, with \emph{scalar rewards}.

In our setting, the agent perceives a \emph{stream of rewards} $r_t^i$, indexed by $i$ at time $t$. To learn the actor neural network parameterized by $\theta$, we are interested in simultaneously maximizing the expected value of all tasks, that is
\begin{align*}
    J(\theta)
    &=
    \E_{\rho^\beta}\Big[
    \sum_i Q_\mu^i(\vs, \mu^i_\theta(\vs))
    \Big],
\end{align*}
where $\mu_\theta^i$ is the actor's policy associated with the $i$-th task, mapping a state vector $\vs$ to a continuous action vector $\va$, and $Q^i$ is the action-value critic associated with this task. The expectation above is taken with respect to $\rho^\beta$, the stationary distribution of some behavior policy $\beta(\va|\vs)$. Note that due to the fact that multiple policies are being learned at once we must necessarily be learning off-policy. The corresponding gradient for the actor is
\begin{align}
    \nabla_\theta J(\theta)
    &
    \approx
    \E_{\rho^\beta}\Big[
    \sum_i 
    \nabla_\theta \mu_\theta^i(\vs)\,
    \nabla_{\va^i} Q_\mu^i(\vs, \va^i)\big|_{\va^i=\mu^i_\theta(\vs)}
    \Big].
\end{align}
The behaviour policy is effectively given by the intentional policies as we will detail shortly. Given an observation $\vs_t$, the behavior policy produces the action vector $\va_t$. In response, the environment returns a \emph{reward vector} $\vr_t$, with one scalar component for each task, and the next state observation $\vs_{t+1}$. The tuple $(\vs_t,\va_t,\vr_t,\vs_{t+1})$ is stored in a replay buffer.   
Note that reward observations do not enter into the gradient estimate as they have instead been captured by the action-value function $Q_\mu$. However, since this quantity is never directly observed, we will instead replace this function with a parameterized critic $Q_w$ which must be trained by an appropriate policy evaluation mechanism. Here we update the critic in order to simultaneously minimize the temporal difference error of all tasks.

To update the critic and actor, we sample a mini-batch of tuples uniformly at random from the replay buffer and perform stochastic gradient descent with respect to both the actor and critic losses. Combining these updates we have,
\begin{align}
    \delta_j^i
    &=
    r_j^i + \gamma Q_{w'}^i(\vs_{j+1}, 
                           \mu^i_{\theta'}(\vs_{j+1}))
    - Q_w^i(\vs_j,\va_j), \label{eq:delta}
    \\
    w
    &\gets
    w + \alpha_\text{critic}
    \sum_j \sum_i \delta_j^i \,\nabla_w Q_w^i(\vs_j, \va_j),
    \\
    \theta
    &\gets
    \theta + \alpha_\text{actor}
    \sum_j \sum_i
    \nabla_\theta \mu_\theta^i(\vs_j)\,
    \nabla_{\va^i} Q_w^i(\vs_j, \va^i)
    \big|_{\va^i=\mu_{\theta}^i(\vs_j)}.
\end{align}
Here $j$ represents the sampled set of indices for the mini-batch. Note that in equation~(\ref{eq:delta}), the same action sampled from the replay buffer, $\va_j$, is used to update every critic $Q^i$. We have also adopted target parameters $\theta'$ and $w'$ which stabilize learning of the critic, as in \citep{lillicrap:2016}. The target parameters are periodically updated to reflect the current values of the optimization parameters. In practice, rather than using fixed learning rates for the actor and critic we instead use dynamic learning rates, e.g.\ via the Adam optimizer \citep{kingma:2015}.

We can now discuss the behavior policy used during training. In this work we use 
\begin{align*}
    \beta(\va|\vs) = \mu_\theta^i(\vs) + Z,
\end{align*}
where $Z$ is a random variable introduced for exploration; following \cite{lillicrap:2016} we utilize temporally correlated noise from an Ornstein-Uhlenbeck process. As before, the index $i$ denotes the intentional task that is being followed. We considered different variations on this process, for example following a single task, selecting a random task to follow at the beginning of each episode, and switching between tasks whenever the current task is successful. However, for clarity of presentation, we focus on having a single intentional policy solving one of the hardest tasks in the physical playroom domain. We will return to this point in Section~\ref{sec:discussion}

\section{Experimental Setup}
\label{sec:experiments}

Our experiments are set in a virtual physical world in which the agent can interact with a variety of objects and obtain rewards by satisfying many procedurally-generated semantic relations between the objects. The interactions among the physical objects involve complex contact forces, which can pose significant challenges for control algorithms.   

Crucially, the agent is physically embodied in this domain and its actions have consistent dynamics throughout the space. Embodiment, object commonality and consistent physics enable the agent to learn features that effectively generalize between different tasks. In this section we will first describe the physical environment in which the agent is situated and then detail the method by which we automatically generate tasks and grounded rewards.

\subsection{The physical playroom domain}

\setlength{\columnsep}{0.75cm}

\begin{wrapfigure}{r}{0.36\textwidth}
    \centering
    \includegraphics[trim={0px 0px 0 0},clip,width=\linewidth]{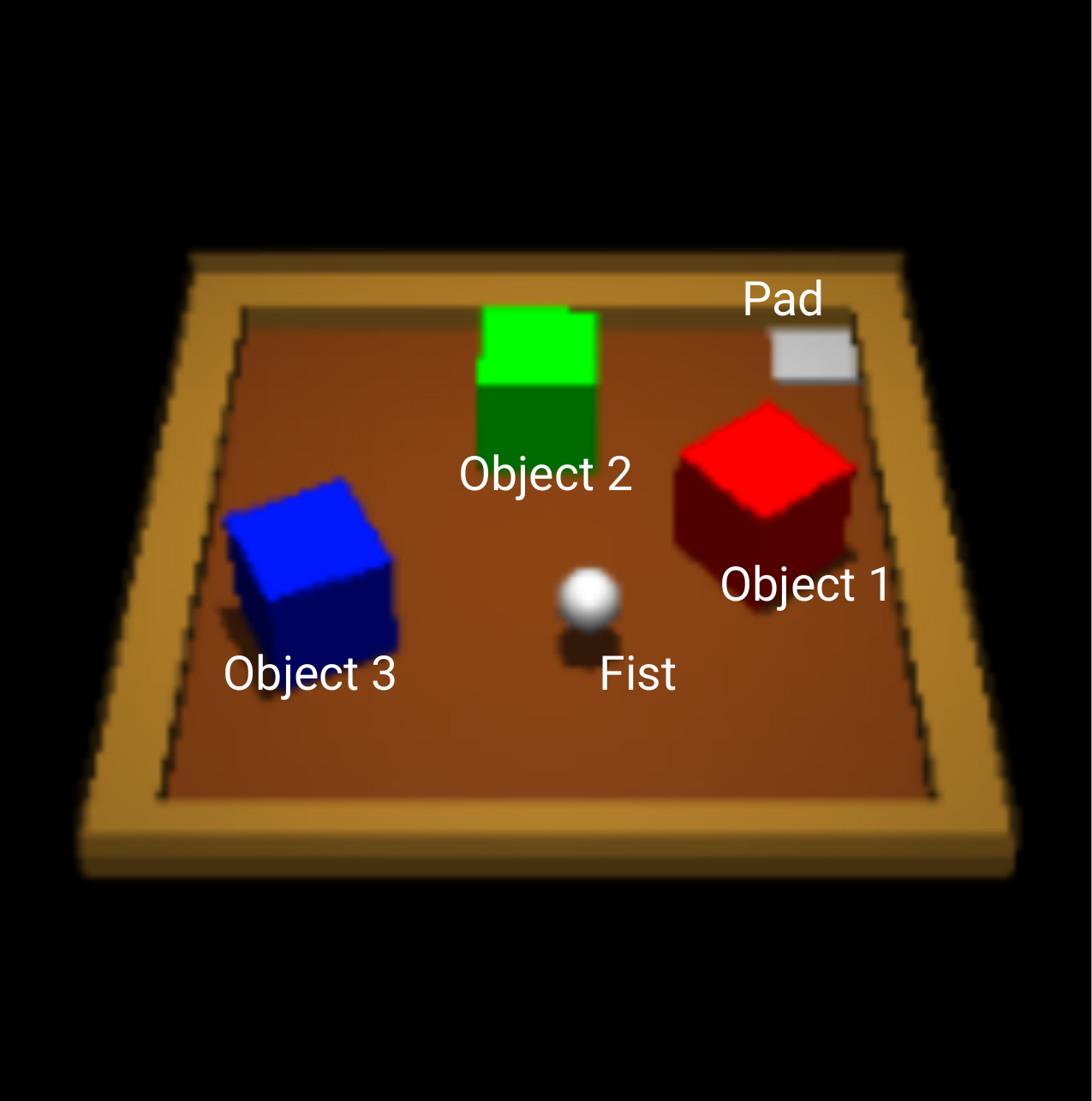}
    \caption{Example of the playroom environment showing three objects and the velocity controlled actuator (fist). Also shown is a goal pad location in the upper right corner.}
    \label{fig:playroom}
\end{wrapfigure}

We implemented a domain using the MuJoCo physics engine \citep{todorov:2012} which consists of a tabletop on which $N$ objects are placed. Each object may have different properties associated with it, for example color, size, friction, density, shape and so on, which allow the objects to be either partially or fully specified. 

The agent is embodied as an actuated ``fist'' whose action space consists of 2-dimensional velocities. In other words, the agent can move its fist through the playroom domain by setting its immediate velocity. The agent is able to interact with other objects and affect their positions only indirectly by way of contact with these objects. Although the fist is only able to move in 2-dimensions the other objects can exhibit more complex behavior based on the speed and angle at which contacts occur.

The playroom consists of a 80 cm$^2$ square arena populated with $N$ objects, each of which is sized to have a diameter of 12 cm.  Figure ~\ref{fig:playroom} shows an example setting of this domain with $N=3$ cubes of different colors. In the figure we can also see the agent's actuated fist, displayed as a white sphere. The fist can be thought of as a directly controllable object, and as a result we will be able to directly express relations between this object and other objects in the scene.

We have also added a border to the environment in order to reduce the occurrence of irrecoverable actions wherein objects are pushed into corners and consequently cannot be retrieved. The border allows the fist to move into the corners but not the blocks. Finally, we will also consider relations between objects and a given goal position. These goals are implemented as immovable goal objects for the purposes of allowing relational expressions; in particular this object is implemented as a \emph{pad} sitting at ground level and hence does not physically interact with any object.
An example of one such object is displayed in Figure~\ref{fig:playroom} as a white pad in the upper right corner.

\subsection{Automatic reward generation with formal language}

Given the above domain we are now interested in generating rewards based on the properties of objects as well as relations between these objects. We will define a number of property functions $p:\calO\times\calS\to\{0,1\}$ where $\calS$ is the set of possible world states. Each property acts as an indicator over objects $o\in\calO$, taking value $1$ when the object satisfies a certain property. It is instructive to think of these properties as defining sets of objects in the scene. 
Often these properties will be independent of the state $s\in\calS$ in which case we can write them simply as $p(o)$.
For example we might define $p_\texttt{blue}(o)$ to be a property of blue objects, which is independent of their location, velocity, etc.


Next we can introduce binary relations between objects which can be thought of as functions of the form $b:\calO\times\calO\times\calS\to\{0,1\}$. For example, we can introduce a ``nearness'' relation which holds when two objects are close to each other, that is when their center points are within some specified distance. By combining properties and relations we can write rewards of the form
\begin{align*}
    r_\texttt{red\_near\_blue}(\vs) 
    &=
    \sum_{o_1, o_2} p_\texttt{red}(o_1) \,p_\texttt{blue}(o_2)
    \,b_\texttt{near}(o_1, o_2, \vs).
\end{align*}
where $o_1$ and $o_2$ represent object identifiers and we are summing over all pairs of objects such that the properties select a particular pair. The above reward represents the task ``bring a red object near a blue object''. If the objects identified in the scene are uniquely identified by their properties this may also be more succinctly written as
\begin{align*}
    r_\texttt{red\_near\_blue}(\vs) 
    &=
    b_\texttt{near}(\texttt{red}, \texttt{blue}, \vs).
\end{align*}
where $\texttt{red}$ and $\texttt{blue}$ are the uniquely identified objects. While properties are frequently independent of the state, relations such as $\texttt{near}$ will depend on this state, specifically the positions of its two input objects $o_1$ and $o_2$. 

We can generate many different rewards by logically combining these operations. In the experiments that follow we will make use of the following atoms:
\begin{enumerate}
    \item color-based properties: $p_\texttt{red}$, $p_\texttt{blue}$, $p_\texttt{green}$;
    \item properties identifying the fist $p_\texttt{fist}$ and the goal state $p_\texttt{goal}$;
    \item near and far relations: $b_\texttt{near}$, $b_\texttt{far}$, parameterized by a distance parameter $\epsilon$ such that whenever the distance between two blocks is less than $\epsilon$ they are considered to be near;
    \item directional relations: $b_\texttt{north}$, $b_\texttt{east}$, $b_\texttt{south}$, $b_\texttt{west}$ such that $b_\texttt{north}(a, b)$ is 1 whenever object $a$ is north of object $b$.
\end{enumerate}
Finally, more complicated rewards can be expressed by and-ing these relations. A ``gather to pad'' task which collects blocks of the three different colors to a goal pad location can be written as
\begin{align*}
    r_\texttt{gather\_to\_pad}(\vs)
    &=
    b_\texttt{near}(\texttt{red}, \texttt{pad}, \vs)\,
    b_\texttt{near}(\texttt{blue}, \texttt{pad}, \vs)\,
    b_\texttt{near}(\texttt{green}, \texttt{pad}, \vs).
\end{align*}
One can also simply gather blocks together or move them far apart from each other, without having to specify a pad or coordinates.

\section{Results}
\label{sec:results}

\begin{figure}[t!]
\begin{center}
    \includegraphics[clip,width=0.49\textwidth]{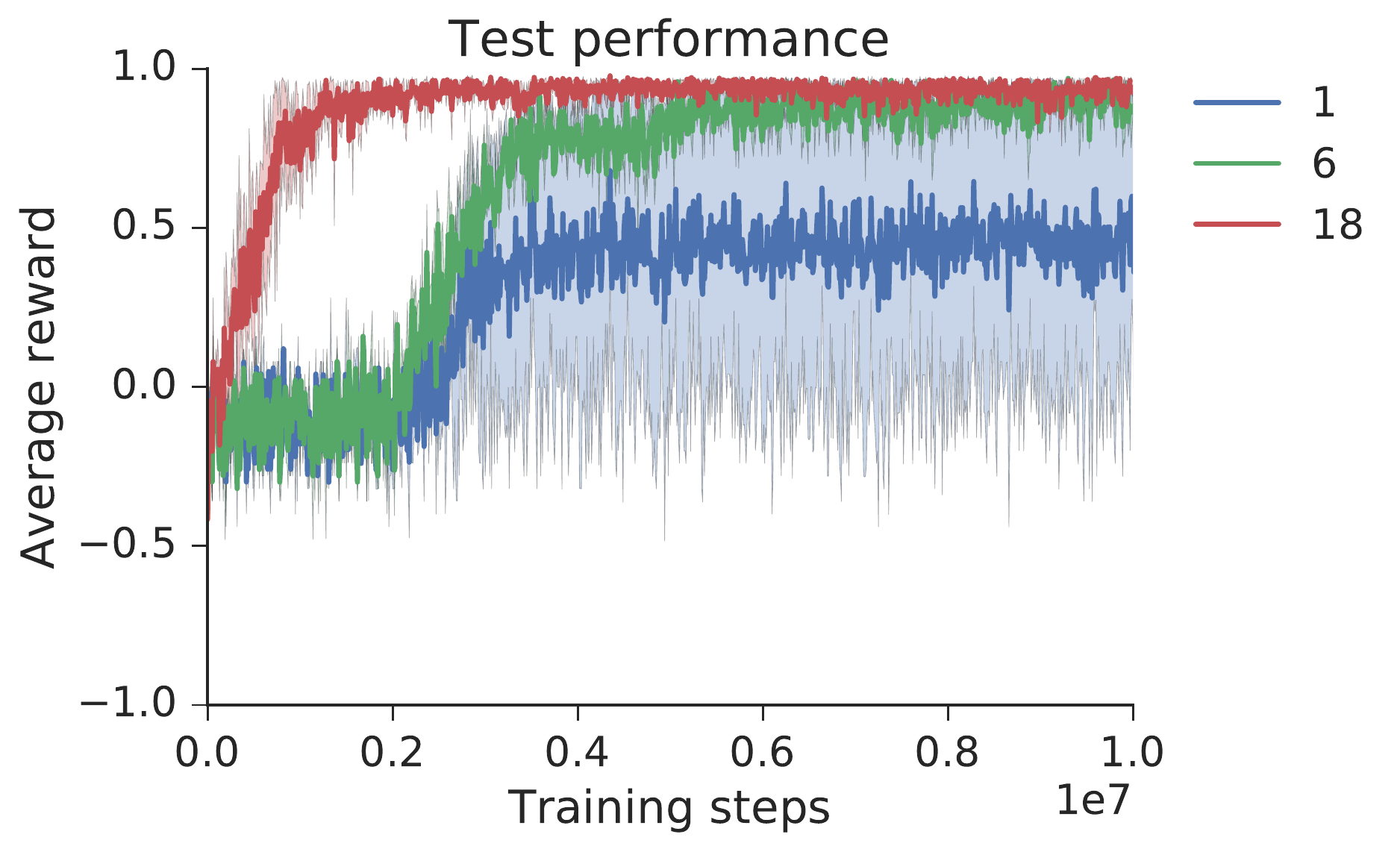}
    \includegraphics[clip,width=0.49\textwidth]{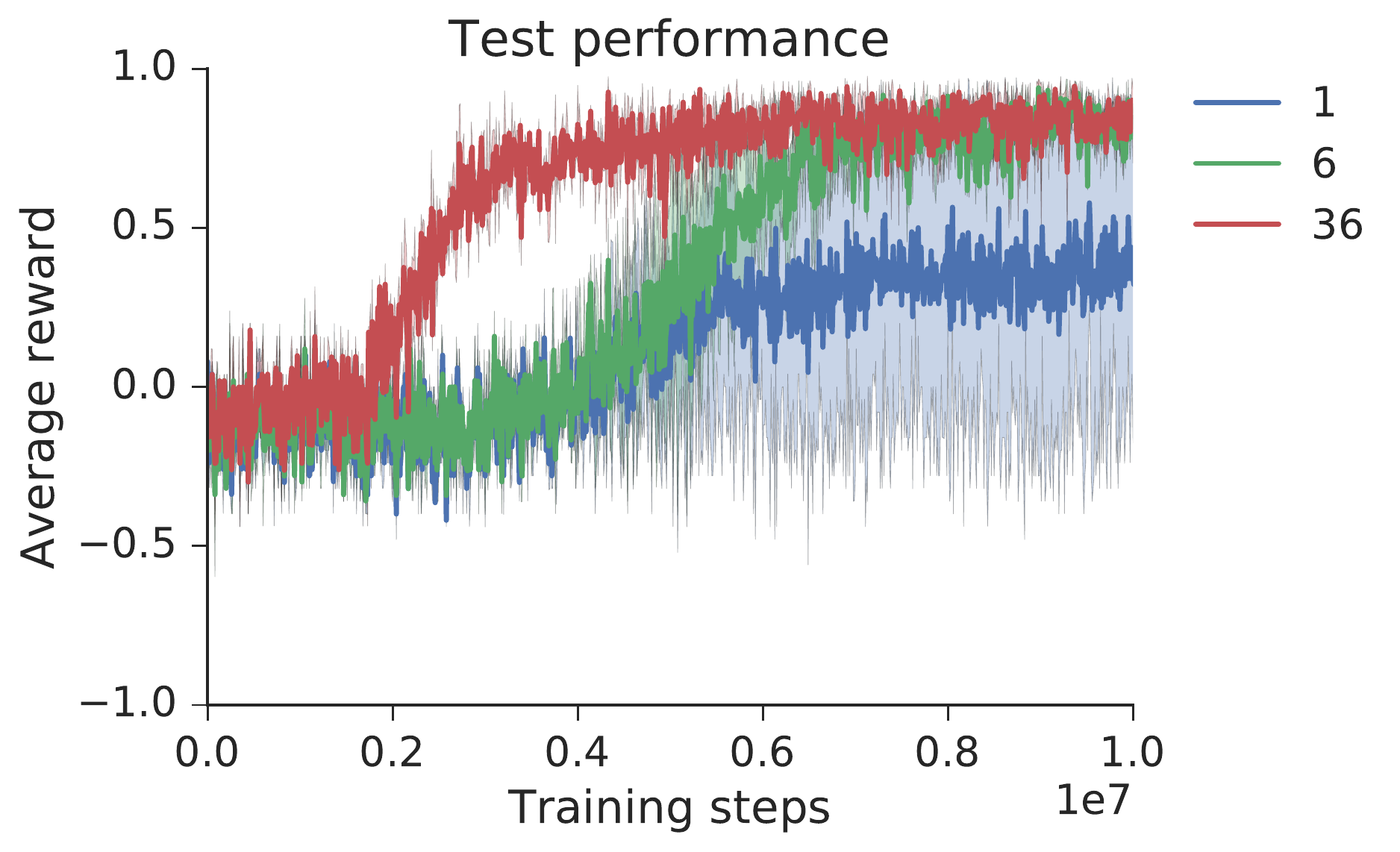}
\end{center}
    \caption{{\bf Left}: Test performance for the task of gathering two blocks together in the playroom, using a varying number of additional tasks. {\bf Right}: What happens when an extra block is added to the environment, causing significant physical interference. In both cases, the more tasks the IU agent solves simultaneously, the faster it learns the intentional task.}
        \label{fig:two-blocks}
\end{figure}

The agent observes the relative position of the fist actuator to each object in the scene. Actions are taken by moving the fist in 2-dimensions, which involves setting the velocities of the actuator. We followed the training protocol described in Appendix A of the DDPG paper of \citet{lillicrap:2016}. The actor and critic multi-layer perceptrons, shown in Figure 1, have 200 and 400 units in each layer respectively and standard hyperbolic tangent activations.

We first consider tasks involving two uncontrolled objects–––identified by the red and blue properties---as well as the fist actuator. 
Our tasks consist of 6 relations: north, south, east, west, near, and far, between each pair of objects including the fist. There are 3 distinct pairs of objects, which results in 18 total reward functions. We are interested in how the use of extra reward functions affects the ability of our agent to learn a single intentional policy.

Figure~\ref{fig:two-blocks} (left) illustrates the ability of our agent to learn to maximize the single intentional reward $b_\texttt{near}(\texttt{red},\texttt{blue}, \vs)$, when given access to additional signals. We consider three scenarios: no additional unintentional tasks (i.e.\ standard DDPG), all near and far tasks (6), and finally all 18 tasks.
 We see that by considering all 18 tasks simultaneously, the IU agent learns to control the system under a single intentional reward much more rapidly. In these plots the test-time reward is shown, averaged over 50 runs, where the error bounds show the min and max rewards for a given evaluation.

\begin{wrapfigure}{r}{0.55\textwidth}
\vspace{-10mm}
\begin{center}
    \includegraphics[clip,width=0.46\textwidth]{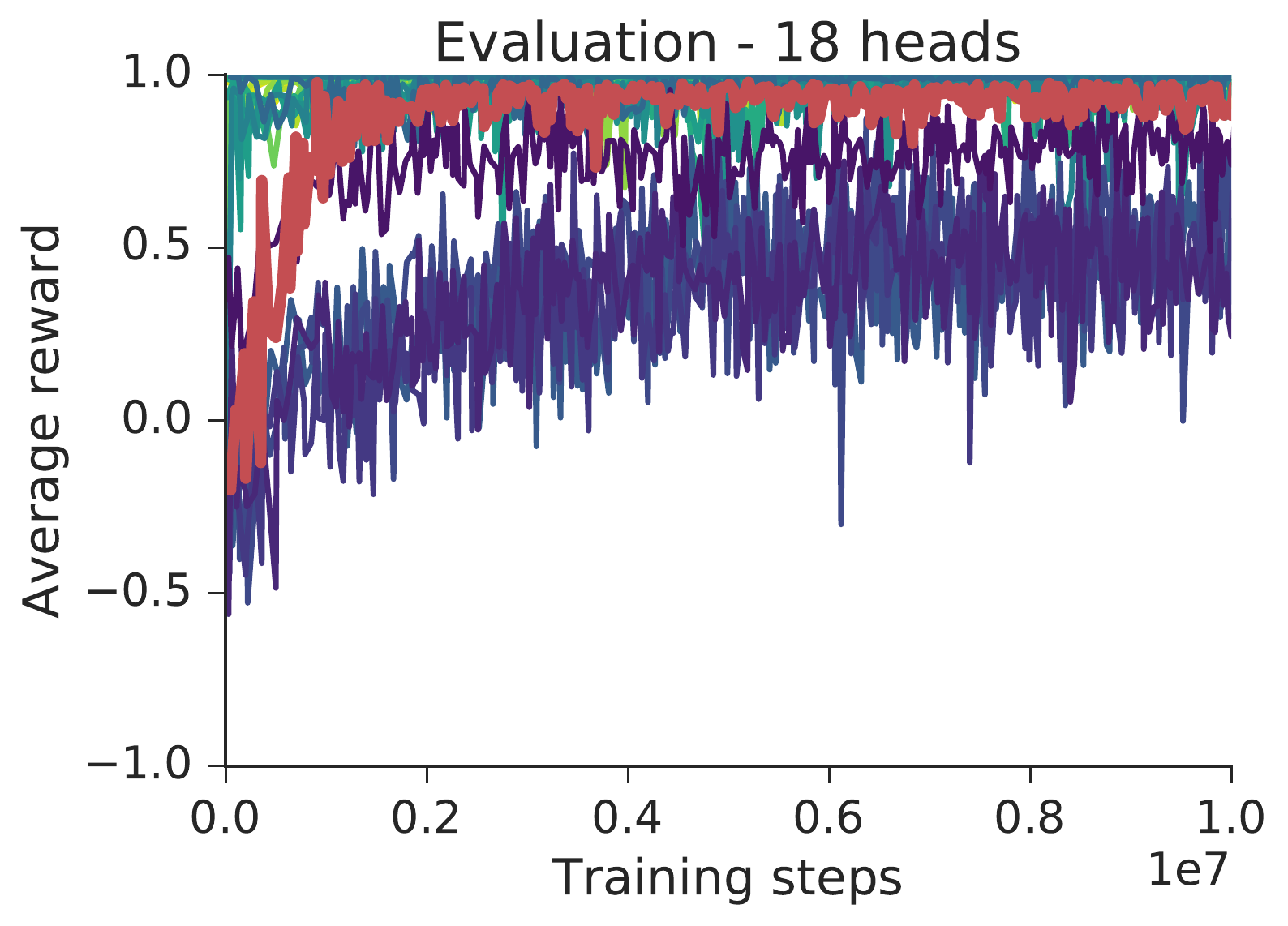}
\end{center}    
\vspace{-2mm}
    \caption{Test performance of the 18 policies 17 of which are inattentional and hence trained off-policy, with the intentional one shown in red, learned simultaneously in the experiment corresponding to Figure 3 (left). All 18 tasks are solved simultaneously.} 
        \label{fig:all-heads}
\vspace{-3mm}
\end{wrapfigure}

Figure~\ref{fig:all-heads} shows the test performance of the 18 policies learned simultaneously. Note that some tasks are easier than others, explaining the different average rewards to which each of the tasks converge. Those tasks which are more difficult require more time to obtain non-zero values and hence converge to an average return less than 1.0. All tasks, however, are able to improve over their initial baseline. 

Next we examine the ability of the agents to learn when an additional (green) cube is added to the playroom. The additional object results in 6 distinct pairs of objects and as a result increases the number of tasks to 36 ($6\times 6$). The results are shown in the right-most plot of Figure~\ref{fig:two-blocks}. We see that although learning is slower, the system is still able to learn to achieve the gathering task despite the distracting block, which can of course cause significant interference. Once again, the more tasks the agent solves the faster it learns. 

\begin{figure}[t!]
\begin{center}
    \belowbaseline[0pt]{
    \includegraphics[clip,width=0.49\textwidth]{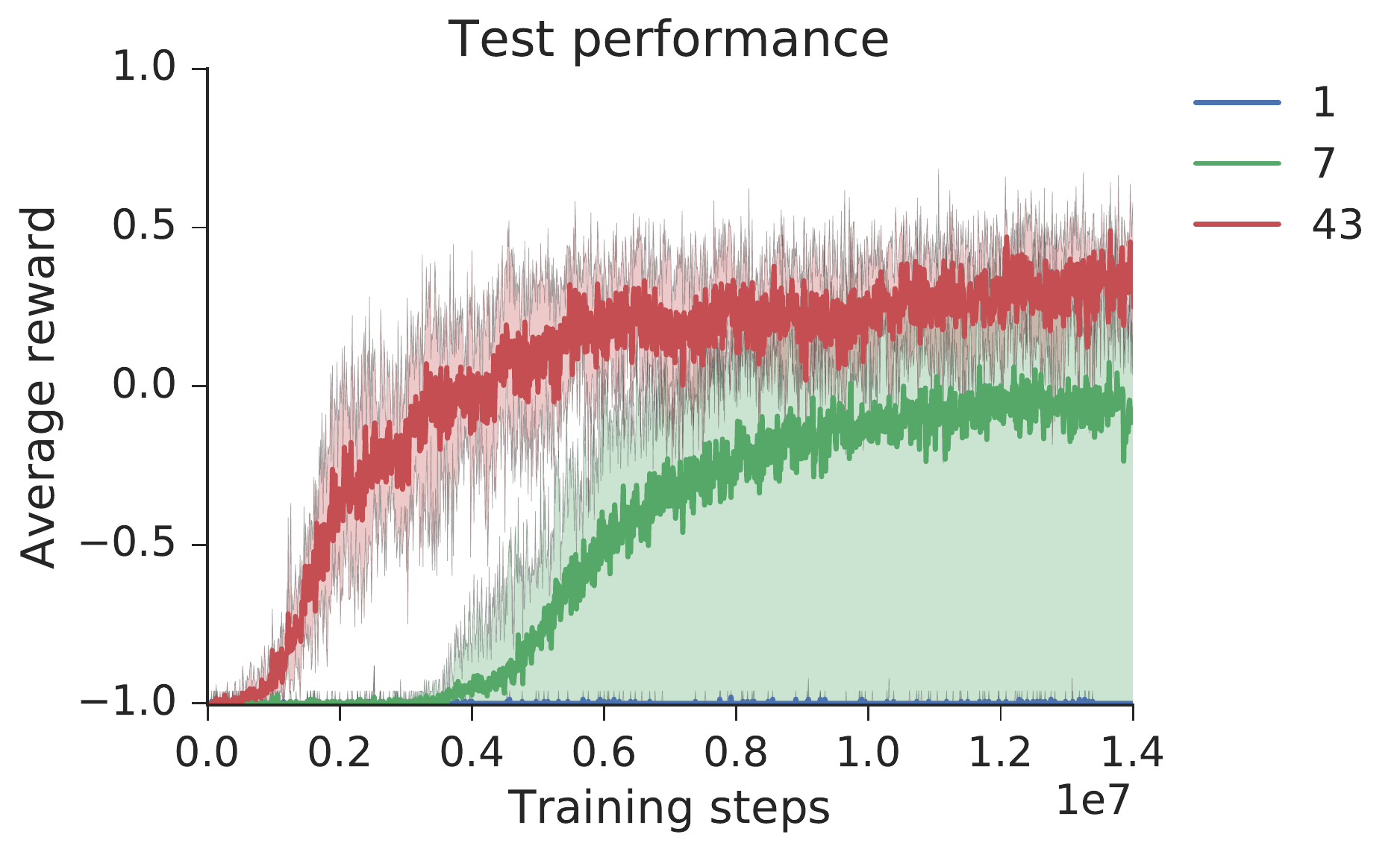}}
    \belowbaseline[0pt]{
    \includegraphics[clip,width=0.49\textwidth]{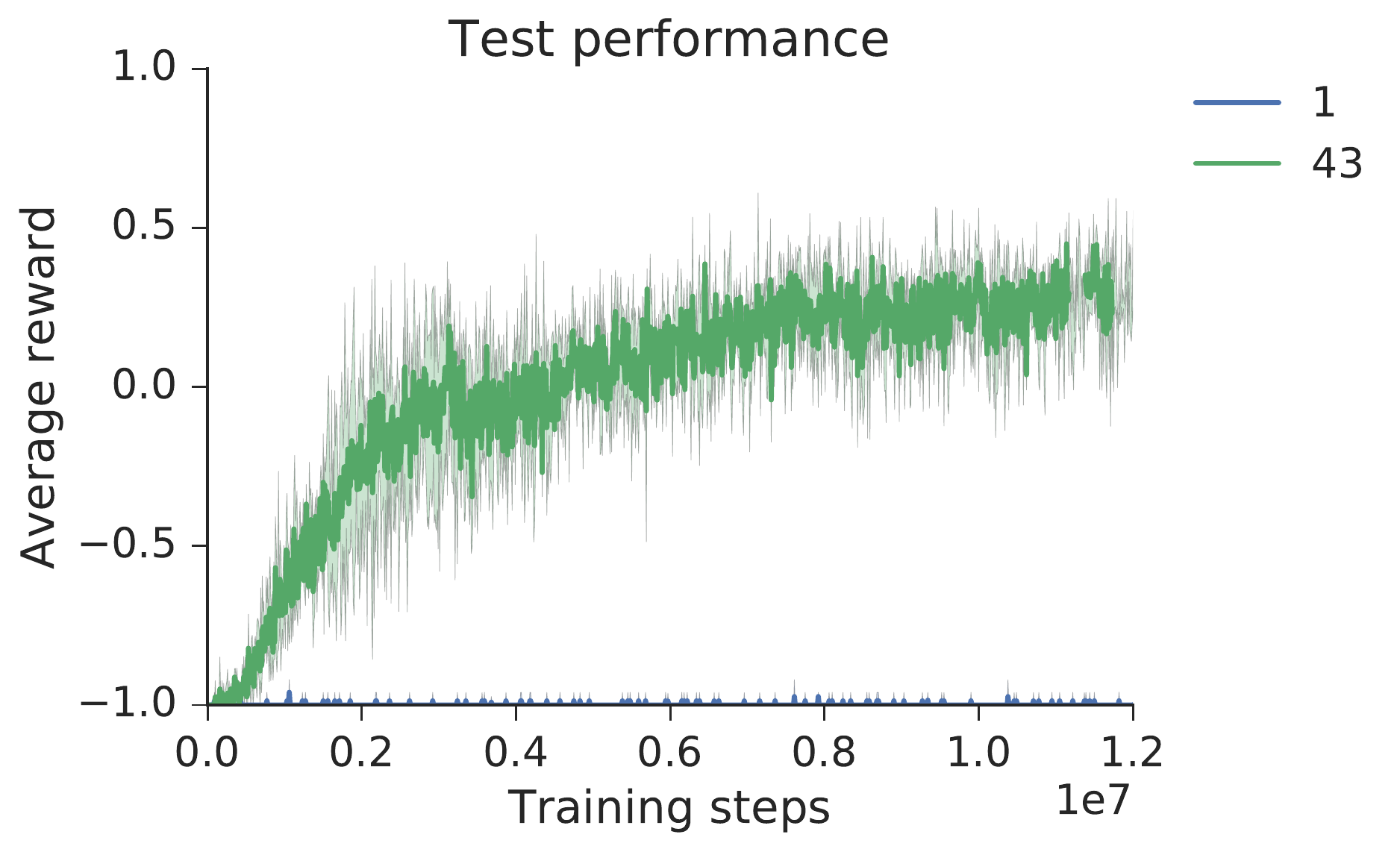}}
 \end{center}   
    \caption{{\bf Left}: Test performance for the task of moving three blocks to the corner pad of the playroom.  {\bf Right}: Same task after increasing the length of the playroom sides by 50\%. When the length is doubled the task can not be learned at all. For these very hard exploration task, the IU agent performs reasonably, while DDPG completely fails.}
        \label{fig:three-blocks}
\end{figure}

In Figure~\ref{fig:three-blocks}, we display the performance of the agent on tasks with three blocks. Here the intentional task involves placing all three blocks in the upper right corner pad. We display the average reward for this task when solving 1, 7, and 43 tasks.  DDPG when following just a single task is incapable of succeeding at this task. However, the intentional unintentional agent succeeds when following 7 and 43 additional rewards. On the right panel of Figure~\ref{fig:three-blocks}, we display the performance of the agent when trained on a larger task space, that is a ground with sides that are 50\% larger in length. We observe that eventually, as the complexity of exploration increases, even the IU agents struggle. However by following 43 tasks the IU agent is still capable of gaining some reward, whereas DDPG completely fails.

\begin{figure}[t!]
    \begin{center}
    \includegraphics[width=0.96\textwidth]{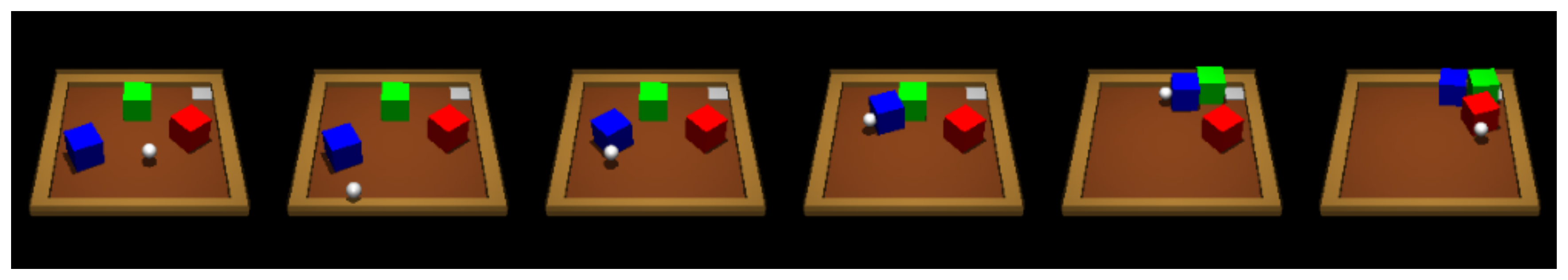}
    \includegraphics[width=0.96\textwidth]{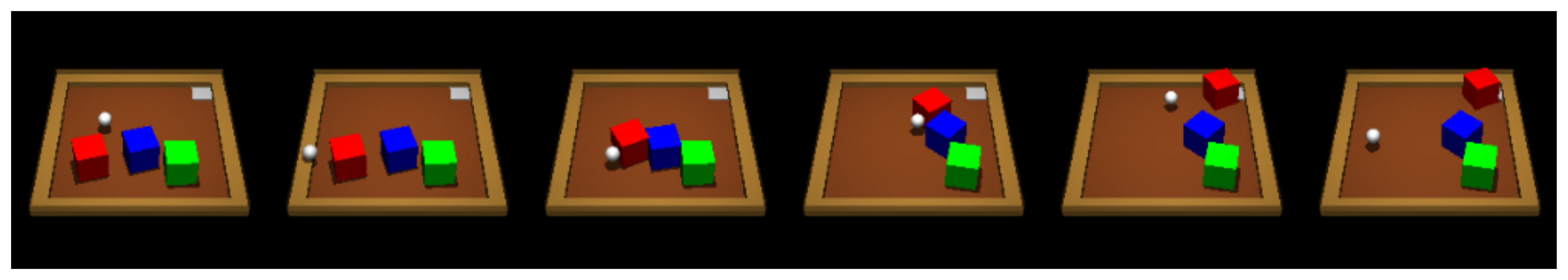}
    \includegraphics[width=0.96\textwidth]{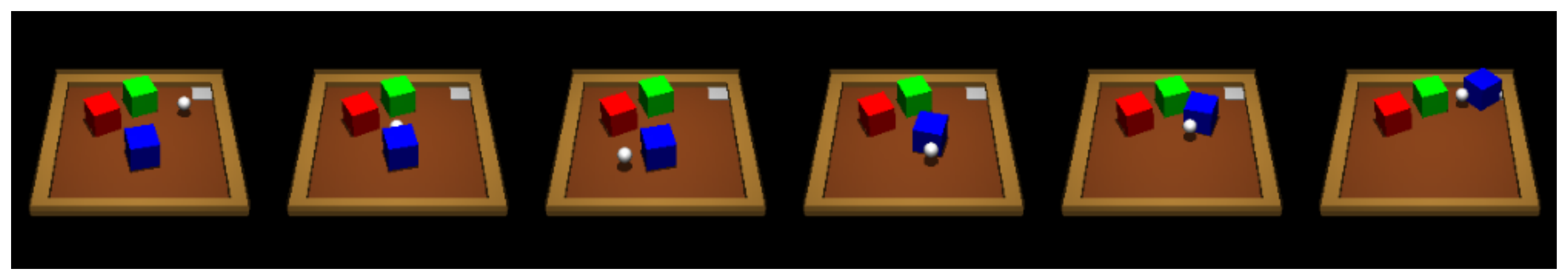}
    
    \includegraphics[width=0.32\textwidth]{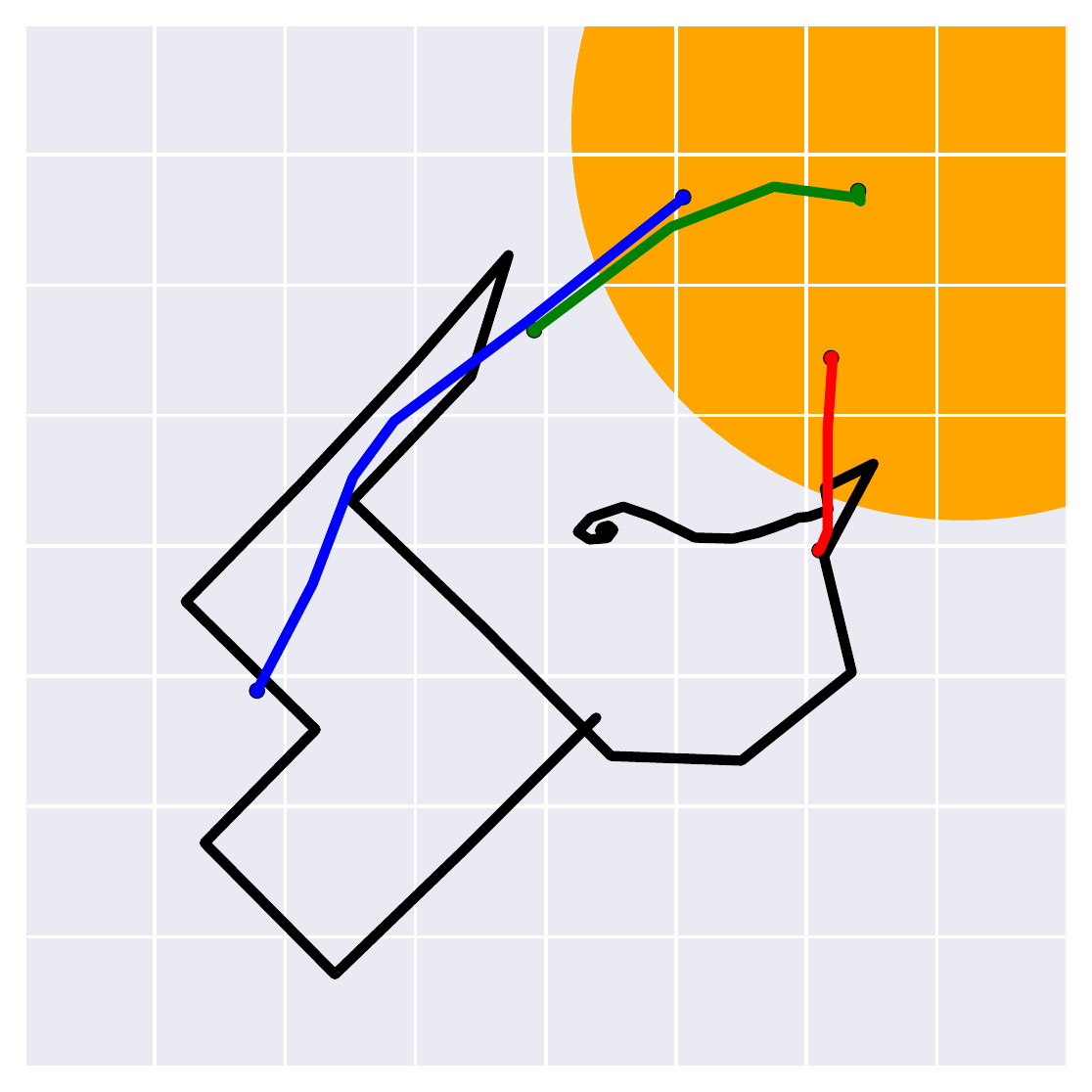}
    \includegraphics[width=0.32\textwidth]{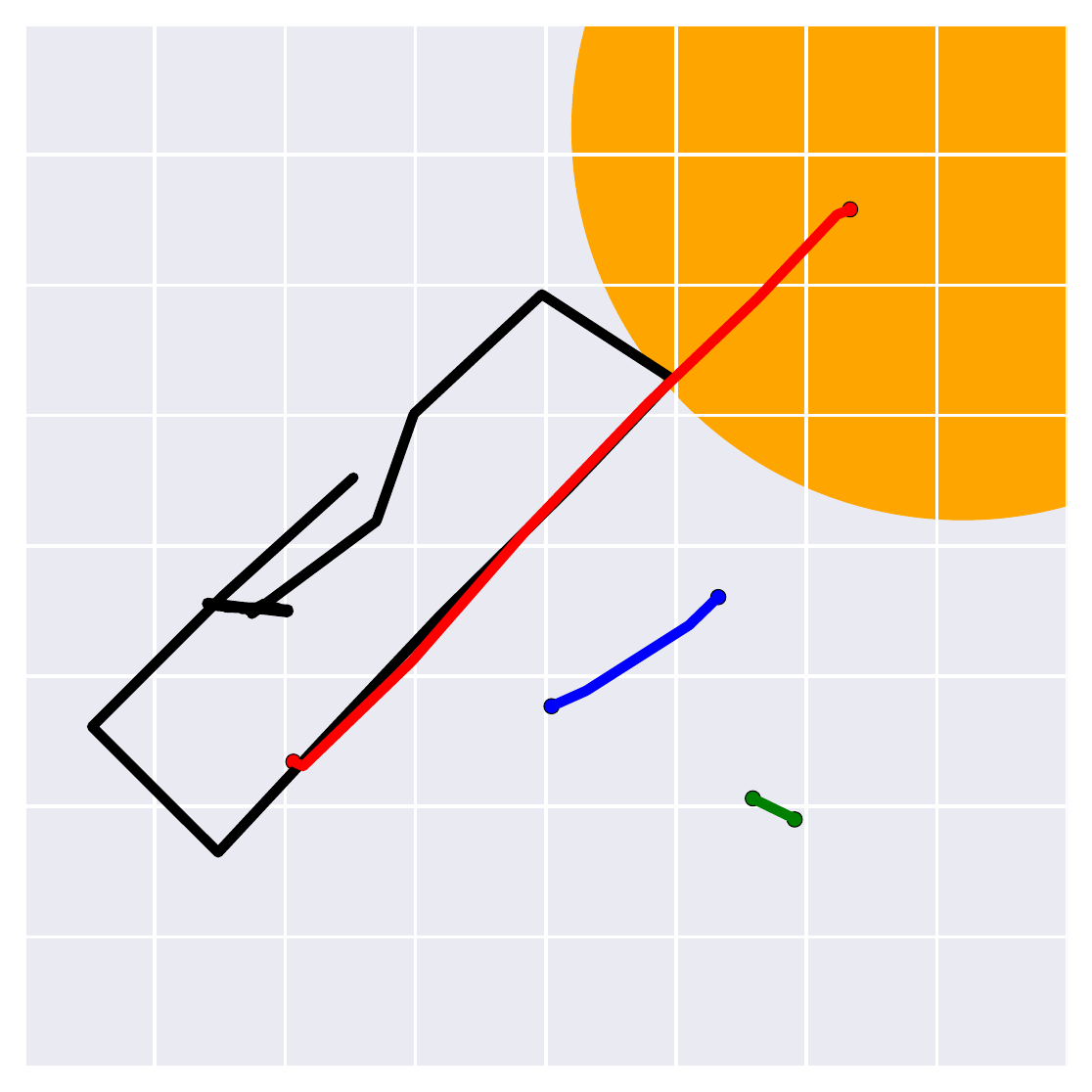}
    \includegraphics[width=0.32\textwidth]{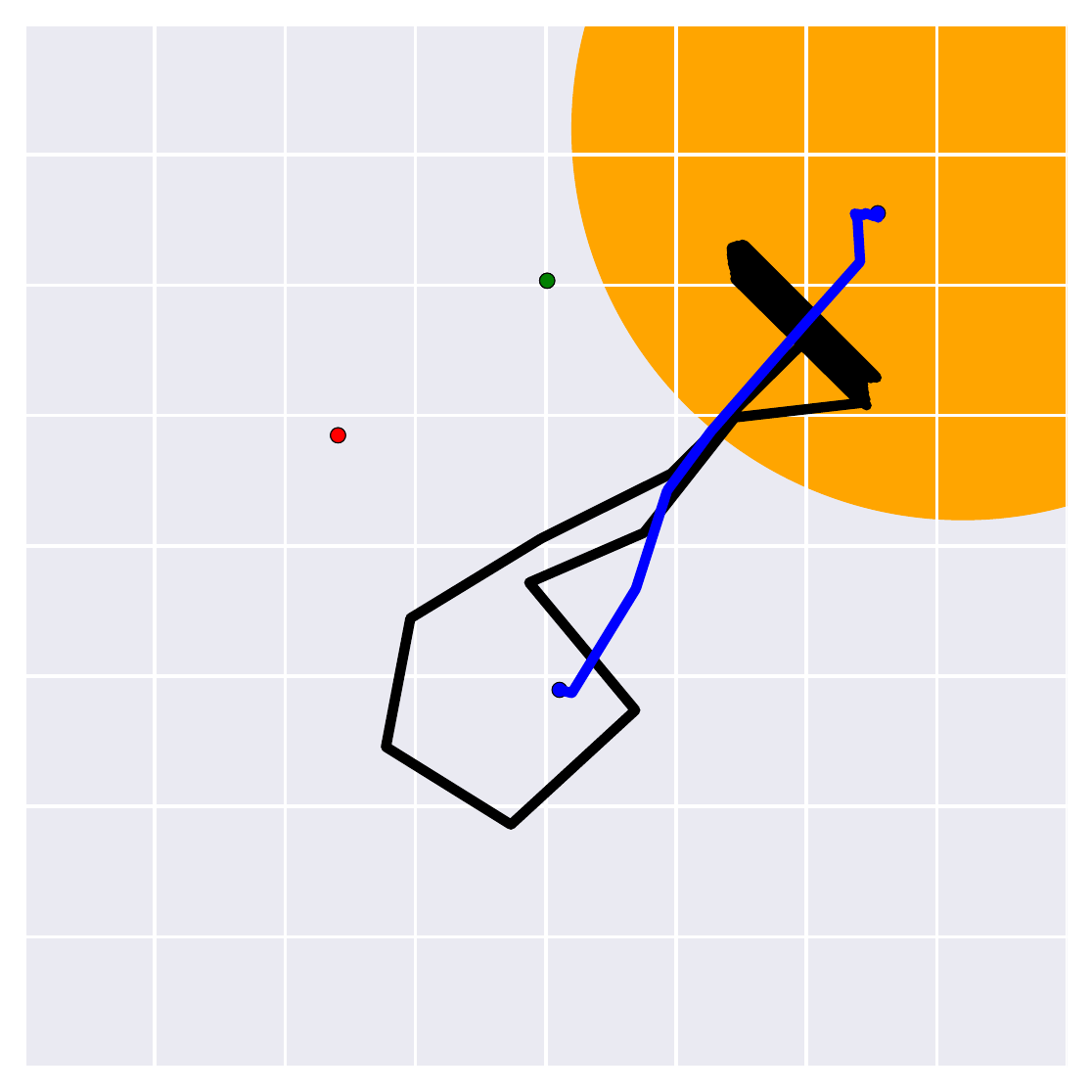}
    \end{center}  
    \vspace{-2mm}
    \caption{IU agent executing three different policies. The top film-strip shows the agent solving the hardest task: gathering all three blocks into the upper right corner. The two strips below this show the agent moving a single block into the corner; red and blue respectively.
    Below these temporal illustrations is a figure illustrating a coordinate-centric view of each of these three tasks. Here the black line denotes the position of the fist as a function of time, and the colored lines represent the positions of each of the three colored blocks. We can see from the first plot that the agent moves each block to reach the goal, shown as an orange semicircle. For the second and third tasks, only the red and blue blocks are acted on respectively, although we can see in the second plot that the agent unintentionally moves the other two blocks slightly.}
    \label{fig:strip}
\end{figure}

Finally, in Figure~\ref{fig:strip} we display frames from example trajectories for various policies learned in the 3-block domain. Shown below these trajectories are illustrations of the object positions over time. The IU agent is able to learn reasonable action sequences which move the objects to the goal pad.

\section{Discussion}
\label{sec:discussion}

Our experiments demonstrate that when acting according to the policy associated with one of the hardest tasks, we are able to learn all other tasks off-policy. The results for the playroom domain also showed that by increasing the number of tasks, all actors and critics learn faster. In fact, in some settings, learning with many goals was essential to solve hard many-body control tasks.

It is worth entertaining what happens when instead of following the policy associated with one of the hardest tasks, we follow a different behaviour strategy. For instance, what happens if we act according to a uniformly random mixture of actors? Is it advantageous to switch policies each time a policy succeeds in encountering a reward? Is it sound to choose behavior actors adaptively and in accordance with the critics' values? 

The above questions are related to the problem of learning curricula. We conducted experiments to explore each of these questions, but found that the naive strategy of choosing the one of the hardest tasks for the behaviour policy works best. There are reasons for this. The behaviour policy determines what information is written to replay memory. These memories in turn are used to update the remaining actor-critics off-policy by gradient descent. When acting according to policies solving simple tasks, the replay memory ends up consisting mostly of experiences associated with the simple tasks, and consequently, the IU agent fails to explore. Populating the memory with rich experiences is essential for learning to solve tasks that involve more exploration. 

Our experimental setup was restricted in the sense that the tasks had a nested structure. That is, approaching a block is needed to bring two blocks together, which in turn is needed to bring three blocks together, and so on. Our conclusions regarding curricula might be different in the absence of this nesting of tasks. This should be studied in future work.

It is not difficult to make the hardest task sufficiently difficult, from an exploration perspective, that even the IU agent fails. For example, if we increase the size of the playground and ask the actuator to bring ten blocks together, all existing control agents are likely to fail. The solution to this hard exploration problem appears to be one of task decomposition, either through hierarchical reinforcement learning or an understanding of objects and relations.  

In this paper we focused on tasks that are sufficiently hard that not even popular continuous control algorithms such as DDPG can solve them. For this reason we obviated other experimental factors, such as perception from pixels, articulated bodies with more degrees of freedom, and  diverse sets of objects. Each of these constitute important challenges that should be addressed in future research.

This work also did not touch upon the topic of policy re-use. That is, once the various policies are learned how do we construct other controllers that can harness these policies to solve new tasks. A simple way to achieve this is to train agents that combine all the policies, either via weighted combinations or more sophisticated deep networks, to construct new policies. Related to this, our work did not address the problem of pruning irrelevant policies. 

In regard to hierarchical RL agents, it would be worthwhile investigating architectures where intrinsic motivation provides the ultimate reward, and is used to guide the automatic online addition or pruning of actors in the IU agent architecture.

\vspace{-2mm}
\section{Conclusions}
\label{sec:conclusions}
\vspace{-3mm}

We present in this work a novel architecture for learning several tasks at once. 
We also propose a flexible way of generating tasks in simulated physical environments. While these environments pose significant challenges because of complex contact forces among the objects, continuous action spaces, and difficult exploration, the body, objects and physical laws are shared among the tasks. In these domains, the more tasks, the faster the IU agents learn. In fact, they can learn to solve complex tasks where popular DDPG agents completely fail.



\clearpage


\bibliography{multihead}

\begin{thebibliography}{34}
\providecommand{\natexlab}[1]{#1}
\providecommand{\url}[1]{\texttt{#1}}
\expandafter\ifx\csname urlstyle\endcsname\relax
  \providecommand{\doi}[1]{doi: #1}\else
  \providecommand{\doi}{doi: \begingroup \urlstyle{rm}\Url}\fi

\bibitem[Thelen(1979)]{thelen:1979}
E.~Thelen.
\newblock Rhythmical stereotypies in normal human infants.
\newblock \emph{Animal Behaviour}, 27:\penalty0 699--715, 1979.

\bibitem[Adolph and Robinson(2015)]{adolph:2015}
K.~E. Adolph and S.~R. Robinson.
\newblock Motor development.
\newblock In R.~M. Lerner, editor, \emph{Handbook of Child Psychology and
  Developmental Science}, volume~2, pages 114--157. John Wiley and Sons, 2015.

\bibitem[Silver et~al.(2014)Silver, Lever, Heess, Degris, Wierstra, and
  Riedmiller]{silver:2014}
D.~Silver, G.~Lever, N.~Heess, T.~Degris, D.~Wierstra, and M.~Riedmiller.
\newblock Deterministic policy gradient algorithms.
\newblock In \emph{International Conference on Machine Learning}, 2014.

\bibitem[Lillicrap et~al.(2016)Lillicrap, Hunt, Pritzel, Heess, Erez, Tassa,
  Silver, and Wierstra]{lillicrap:2016}
T.~P. Lillicrap, J.~J. Hunt, A.~Pritzel, N.~Heess, T.~Erez, Y.~Tassa,
  D.~Silver, and D.~Wierstra.
\newblock Continuous control with deep reinforcement learning.
\newblock In \emph{International Conference on Learning Representations}, 2016.

\bibitem[Jaderberg et~al.(2017)Jaderberg, Mnih, Czarnecki, Schaul, Leibo,
  Silver, and Kavukcuoglu]{Jaderberg:2017}
M.~Jaderberg, V.~Mnih, W.~M. Czarnecki, T.~Schaul, J.~Z. Leibo, D.~Silver, and
  K.~Kavukcuoglu.
\newblock Reinforcement learning with unsupervised auxiliary tasks.
\newblock In \emph{International Conference on Learning Representations}, 2017.

\bibitem[Mnih et~al.(2016)Mnih, Badia, Mirza, Graves, Lillicrap, Harley,
  Silver, and Kavukcuoglu]{mnih:2016}
V.~Mnih, A.~P. Badia, M.~Mirza, A.~Graves, T.~Lillicrap, T.~Harley, D.~Silver,
  and K.~Kavukcuoglu.
\newblock Asynchronous methods for deep reinforcement learning.
\newblock In \emph{International Conference on Machine Learning}, pages
  1928--1937, 2016.

\bibitem[Sutton et~al.(2011)Sutton, Modayil, Delp, Degris, Pilarski, White, and
  Precup]{sutton:2011}
R.~S. Sutton, J.~Modayil, M.~Delp, T.~Degris, P.~M. Pilarski, A.~White, and
  D.~Precup.
\newblock Horde: A scalable real-time architecture for learning knowledge from
  unsupervised sensorimotor interaction.
\newblock In \emph{International Conference on Autonomous Agents and Multiagent
  Systems}, pages 761--768, 2011.

\bibitem[Schaul et~al.(2015)Schaul, Horgan, Gregor, and Silver]{Schaul:2015}
T.~Schaul, D.~Horgan, K.~Gregor, and D.~Silver.
\newblock Universal value function approximators.
\newblock In \emph{International Conference on Machine Learning}, pages
  1312--1320, 2015.

\bibitem[Dietterich(1998)]{Dietterich:1998}
T.~G. Dietterich.
\newblock The {MAXQ} method for hierarchical reinforcement learning.
\newblock In \emph{International Conference on Machine Learning}, 1998.

\bibitem[Sutton et~al.(1998)Sutton, Precup, and Singh]{Sutton:1998}
R.~S. Sutton, D.~Precup, and S.~P. Singh.
\newblock Intra-option learning about temporally abstract actions.
\newblock In \emph{International Conference on Machine Learning}, pages
  556--564, 1998.

\bibitem[Kulkarni et~al.(2016)Kulkarni, Narasimhan, Saeedi, and
  Tenenbaum]{Kulkarni:2016}
T.~D. Kulkarni, K.~Narasimhan, A.~Saeedi, and J.~Tenenbaum.
\newblock Hierarchical deep reinforcement learning: Integrating temporal
  abstraction and intrinsic motivation.
\newblock In \emph{Advances in Neural Information Processing Systems}, pages
  3675--3683, 2016.

\bibitem[van Seijen et~al.(2017)van Seijen, Fatemi, Romoff, Laroche, Barnes,
  and Tsang]{vanSeijen2017}
H.~van Seijen, M.~Fatemi, J.~Romoff, R.~Laroche, T.~Barnes, and J.~Tsang.
\newblock Hybrid reward architecture for reinforcement learning.
\newblock Technical report, Microsoft Maluuba, 2017.

\bibitem[Russell and Zimdars(2003)]{russell2003}
S.~Russell and A.~Zimdars.
\newblock Q-decomposition for reinforcement learning agents.
\newblock In \emph{International Conference on Machine Learning}, 2003.

\bibitem[Mnih et~al.(2015)Mnih, Kavukcuoglu, Silver, Rusu, Veness, Bellemare,
  Graves, Riedmiller, Fidjeland, Ostrovski, Petersen, Beattie, Sadik,
  Antonoglou, King, Kumaran, Wierstra, Legg, and Hassabis]{Mnih:2015}
V.~Mnih, K.~Kavukcuoglu, D.~Silver, A.~A. Rusu, J.~Veness, M.~G. Bellemare,
  A.~Graves, M.~Riedmiller, A.~K. Fidjeland, G.~Ostrovski, S.~Petersen,
  C.~Beattie, A.~Sadik, I.~Antonoglou, H.~King, D.~Kumaran, D.~Wierstra,
  S.~Legg, and D.~Hassabis.
\newblock Human-level control through deep reinforcement learning.
\newblock \emph{Nature}, 518\penalty0 (7540):\penalty0 529--533, 2015.

\bibitem[Roijers et~al.(2013)Roijers, Vamplew, Whiteson, and
  Dazeley]{Roijers:2013}
D.~M. Roijers, P.~Vamplew, S.~Whiteson, and R.~Dazeley.
\newblock A survey of multi-objective sequential decision-making.
\newblock \emph{J. Artif. Int. Res.}, 48\penalty0 (1):\penalty0 67--113, 2013.

\bibitem[Dosovitskiy and Koltun(2017)]{Dosovitskiy:2017}
A.~Dosovitskiy and V.~Koltun.
\newblock Learning to act by predicting the future.
\newblock 2017.

\bibitem[Li et~al.(2015)Li, Li, Gao, He, Chen, Deng, and He]{li:2015}
X.~Li, L.~Li, J.~Gao, X.~He, J.~Chen, L.~Deng, and J.~He.
\newblock Recurrent reinforcement learning: a hybrid approach.
\newblock \emph{Preprint arXiv:1509.03044}, 2015.

\bibitem[Lample and Chaplot(2016)]{lample:2016}
G.~Lample and D.~S. Chaplot.
\newblock Playing {FPS} games with deep reinforcement learning.
\newblock \emph{Preprint arXiv:1609.05521}, 2016.

\bibitem[Bengio et~al.(2009)Bengio, Louradour, Collobert, and
  Weston]{Bengio:2009}
Y.~Bengio, J.~Louradour, R.~Collobert, and J.~Weston.
\newblock Curriculum learning.
\newblock In \emph{International Conference on Machine Learning}, pages 41--48,
  2009.

\bibitem[Zaremba and Sutskever(2014)]{Zaremba:2014}
W.~Zaremba and I.~Sutskever.
\newblock Learning to execute.
\newblock \emph{Preprint arXiv:1410.4615}, 2014.

\bibitem[Reed and de~Freitas(2016)]{reed:2015}
S.~Reed and N.~de~Freitas.
\newblock Neural programmer-interpreters.
\newblock 2016.

\bibitem[Tsvetkov et~al.(2016)Tsvetkov, Faruqui, Ling, MacWhinney, and
  Dyer]{Tsvetkov:2016}
Y.~Tsvetkov, M.~Faruqui, W.~Ling, B.~MacWhinney, and C.~Dyer.
\newblock Learning the curriculum with {Bayesian} optimization for
  task-specific word representation learning.
\newblock In \emph{Association for Computational Linguistics}, 2016.

\bibitem[Graves et~al.(2017)Graves, Bellemare, Menick, Munos, and
  Kavukcuoglu]{Graves:2017}
A.~Graves, M.~G. Bellemare, J.~Menick, R.~Munos, and K.~Kavukcuoglu.
\newblock Automated curriculum learning for neural networks.
\newblock \emph{Preprint arXiv:1704.03003}, 2017.

\bibitem[Sparling et~al.(1999)Sparling, Van~Tol, and Chescheir]{sparling:1999}
J.~W. Sparling, J.~Van~Tol, and N.~C. Chescheir.
\newblock Fetal and neonatal hand movement.
\newblock \emph{Physical Therapy}, 79\penalty0 (1):\penalty0 24, 1999.

\bibitem[Blumberg et~al.(2013)Blumberg, Marques, and Iida]{blumberg:2013}
M.~S. Blumberg, H.~G. Marques, and F.~Iida.
\newblock Twitching in sensorimotor development from sleeping rats to robots.
\newblock \emph{Current Biology}, 23\penalty0 (12):\penalty0 R532--R537, 2013.

\bibitem[Lungarella et~al.(2003)Lungarella, Metta, Pfeifer, and
  Sandini]{lungarella:2003}
M.~Lungarella, G.~Metta, R.~Pfeifer, and G.~Sandini.
\newblock Developmental robotics: a survey.
\newblock \emph{Connection Science}, 15\penalty0 (4):\penalty0 151--190, 2003.

\bibitem[Brooks(1991)]{brooks:1991}
R.~A. Brooks.
\newblock Intelligence without representation.
\newblock \emph{Artificial intelligence}, 47\penalty0 (1-3):\penalty0 139--159,
  1991.

\bibitem[Littman et~al.(2017)Littman, Topcu, Fu, Isbell, Wen, and
  MacGlashan]{Littman2017a}
M.~L. Littman, U.~Topcu, J.~Fu, C.~Isbell, M.~Wen, and J.~MacGlashan.
\newblock Environment-independent task specifications via {GLTL}.
\newblock Technical report, Brown University, 2017.

\bibitem[Yu et~al.(2017)Yu, Zhang, and Xu]{Yu2017}
H.~Yu, H.~Zhang, and W.~Xu.
\newblock A deep compositional framework for human-like language acquisition in
  virtual environment.
\newblock Technical report, Baidu Research, 2017.

\bibitem[Hermann et~al.(2017)Hermann, Hill, Green, Wang, Faulkner, Soyer,
  Szepesvari, Czarnecki, Jaderberg, Teplyashin, Wainwright, Apps, Hassabis, and
  Blunsom]{Hermann2017}
K.~M. Hermann, F.~Hill, S.~Green, F.~Wang, R.~Faulkner, H.~Soyer,
  D.~Szepesvari, W.~Czarnecki, M.~Jaderberg, D.~Teplyashin, M.~Wainwright,
  C.~Apps, D.~Hassabis, and P.~Blunsom.
\newblock Grounded language learning in a simulated {3D} world.
\newblock Technical report, DeepMind, 2017.

\bibitem[Denil et~al.(2017)Denil, Colmenarejo, Cabi, Saxton, and
  de~Freitas]{Denil2017}
M.~Denil, S.~G. Colmenarejo, S.~Cabi, D.~Saxton, and N.~de~Freitas.
\newblock Programmable agents.
\newblock Technical report, DeepMind, 2017.

\bibitem[Sutton et~al.(1999)Sutton, McAllester, Singh, and
  Mansour]{sutton:1999}
R.~S. Sutton, D.~A. McAllester, S.~P. Singh, and Y.~Mansour.
\newblock Policy gradient methods for reinforcement learning with function
  approximation.
\newblock In \emph{Advances in Neural Information Processing Systems}, 1999.

\bibitem[Kingma and Ba(2015)]{kingma:2015}
D.~Kingma and J.~Ba.
\newblock Adam: A method for stochastic optimization.
\newblock In \emph{International Conference on Learning Representations}, 2015.

\bibitem[Todorov et~al.(2012)Todorov, Erez, and Tassa]{todorov:2012}
E.~Todorov, T.~Erez, and Y.~Tassa.
\newblock Mujoco: A physics engine for model-based control.
\newblock In \emph{International Conference on Intelligent Robots and Systems},
  2012.

\end{thebibliography}

\end{document}